\pdfoutput=1

\documentclass[11pt]{article}

\usepackage{EMNLP2023}

\usepackage{times}
\usepackage{latexsym}
\usepackage{tabularx}

\usepackage[T1]{fontenc}

\usepackage[utf8]{inputenc}

\usepackage{hyperref}

\usepackage{microtype}

\usepackage{inconsolata}

\usepackage{multirow}
\usepackage{booktabs, colortbl} 
\usepackage[normalem]{ulem}
\useunder{\uline}{\ul}{}
\usepackage{algorithm}
\usepackage{algpseudocode}
\usepackage{graphicx}

\usepackage{url}

\title{End-to-End Evaluation for Low-Latency Simultaneous Speech Translation}

\author{Christian Huber$^1$, Tu Anh Dinh$^1$, Carlos Mullov$^1$, Ngoc Quan Pham$^1$,\\
\textbf{Thai Binh Nguyen$^1$, Fabian Retkowski$^1$, Stefan Constantin$^1$, Enes Yavuz Ugan$^1$,}\\
\textbf{Danni Liu$^1$, Zhaolin Li$^1$, Sai Koneru$^1$, Jan Niehues$^1$ and Alexander Waibel$^{1,2}$}\\
  $^1$Karlsruhe Institute of Technology, Karlsruhe, Germany\\
  \texttt{firstname.lastname@kit.edu}\\
  $^2$Carnegie Mellon University, Pittsburgh PA, USA\\
  \texttt{alexander.waibel@cmu.edu}}

\begin{document}
\maketitle
\begin{abstract}

The challenge of low-latency speech translation has recently draw significant interest in the research community as shown by several publications and shared tasks. Therefore, it is essential to evaluate these different approaches in realistic scenarios. However, currently only specific aspects of the systems are evaluated and often it is not possible to compare different approaches.

In this work, we propose the first framework to perform and evaluate the various aspects of low-latency speech translation under realistic conditions. The evaluation is carried out in an end-to-end fashion. This includes the segmentation of the audio as well as the run-time of the different components.

Secondly, we compare different approaches to low-latency speech translation using this framework. We evaluate models with the option to revise the output as well as methods with fixed output.
Furthermore, we directly compare state-of-the-art cascaded as well as end-to-end systems.
Finally, the framework allows to automatically evaluate the translation quality as well as latency and also provides a web interface to show the low-latency model outputs to the user.

\end{abstract}

\section{Introduction}

In many applications scenarios for speech translation, the quality of the translations is not the only important metric, but it is also essential to provide the translation with a low latency. This is for example the case in translations of presentations or meetings. Therefore, we observe an increasing interest in the field of low-latency speech translations, as shown by numerous published techniques and the organization of a dedicated shared task as part of the International Conference on Spoken Language Translations (IWSLT) \cite{agrawal-etal-2023-findings}.

In order to enable further progress 
in the field as well as a wide adoption of the technique a framework to evaluate different approaches is essential. However, the current evaluation
only considers 
a limited number of aspects or techniques. In contrast, for an overall evaluation of different architectures (end-to-end and cascaded) and presentation style
(revision 
and fixed) a general evaluation framework is needed. This should also consider the computational latency as well as the ability to process several sessions in parallel.

Motivated by this, we present a new framework to apply and evaluate low-latency, simultaneous speech translation. Thereby we focus on a framework that can evaluate the different approaches in as realistic conditions as possible. The system is able to simulate different load conditions as well as compare systems using different design choices. Finally, we also provide a web interface\footnote{\url{https://lecture-translator.kit.edu}}
to present the low-latency model outputs to the user.

\begin{figure}[t]
  \vskip -15pt
  \centering
  \includegraphics[width=1.0\columnwidth]{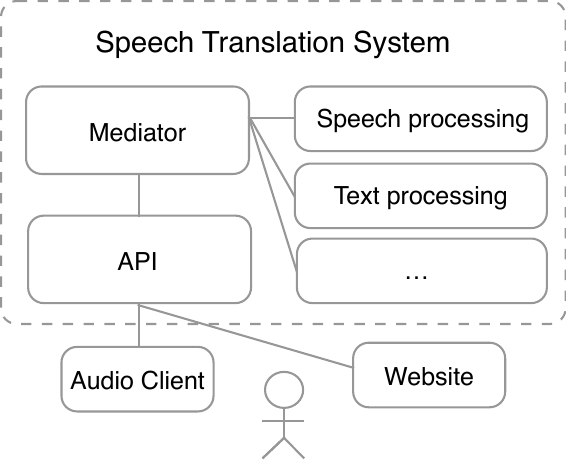}
  \vskip -5pt
  \caption{Framework overview}
  \label{fig:system_overview}
  \vskip -15pt
\end{figure}

The main contributions of our paper are:
\begin{itemize}
    \item A framework\footnote{\url{https://git.scc.kit.edu/isl/lt-middleware/ltpipeline}} for low-latency speech translation with dynamic latency adjustment 
    \item An evaluation setup that allows for assessing the quality and latency of a low-latency scenario in an end-to-end fashion
    \item A comprehensive evaluation of different translation approaches and streaming algorithms
\end{itemize}


In the next section, we describe the overall architecture of the framework.
The two following sections explain the streaming algorithms for the speech and text processing components.
After that, we illustrate how we evaluate our framework and then how the experimental setup looks like.
In Section \ref{sec:results} we present the results. 
Then, we review the related work. At the end we describe the limitations and conclude our work.

\section{Dynamic Framework for low-latency speech translation}


Motivated by previous work \cite{cho13_interspeech}, we use a central mediator that coordinates the interaction of the different components (see Figure \ref{fig:system_overview}). The user sends data to an API component which then sends the data to the mediator.
The mediator forwards all arriving data to the corresponding component(s), e.g., the audio signal from the user to the speech processing component, the resulting transcripts to the text processing component and the output (through the API) to the user. In order to allow a flexible processing, for each session a graph dynamically defines how the data is sent to the different components. We process different requests at each component using the existing streaming framework Kafka\footnote{\url{https://kafka.apache.org}}. 

Each component consists of a middleware and a backend with the processing separated into three steps:

1) Input processing: The middleware
implements the streaming algorithms and can be run on the CPU. It uses the state of the current session to generate requests to the backend.
Other approaches \cite{niehues18_interspeech} repeatedly send requests to the backend for all input messages. This can result in increasing latency if the backend is not able to keep up in high-load situations. In order to minimize this, we enable the middleware to skip intermediate processing steps. 
This is done by 
combining multiple input messages by concatenating audio or text.
Several middleware workers can be run in parallel. We achieve the locality of the state by sticky queues, where a message from the same session is always sent to the same middleware worker. 

2) Backend request: The backend contains the hosted models. It processes the requests without additional state information, is flexible to run on any device and is shared between different
sessions. 
Because of the division in a stateful middleware and a stateless backend, we are able to share the backend and use batching of the requests.

3) Output processing: The output of a backend request is used to send information to the next component(s). Furthermore, the state of the corresponding session is updated.



\begin{figure}[t]
  \centering
  \includegraphics[width=1.0\columnwidth]{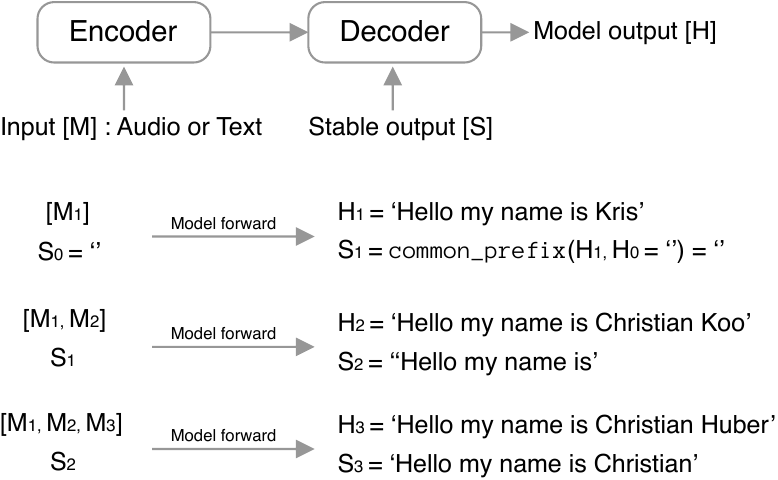}
  \vskip -0pt
  \caption{Stability detection}
  \label{fig:stability_detection}
  \vskip -10pt
\end{figure}

Our framework supports two modes for low-latency speech translation. First, a revision mode \cite{niehues18_interspeech} where the component (Automatic speech recognition (ASR) or machine translation (MT)) can send stable and unstable outputs. Given more context at a later time step, the component can revise the unstable outputs. Second, a fixed mode \cite{liu20s_interspeech,polak2022cuni} where the component is only allowed to send stable output. For fixed mode (and the revision mode of the ASR component), the component needs to perform a stability detection (see Sections \ref{sec:LowLatencySpeechProcessing} and \ref{sec:LowLatencyTextProcessing} and Figure \ref{fig:stability_detection}), i.e., determine which parts of the output should be considered stable.
Note that for our streaming algorithms the backend models need to support prefix decoding, i.e., one can send a prefix which is then forced in the output.

Our framework is easily extendable by deploying additional backend models for different languages, adding new streaming algorithms in the middleware or adding custom components (e.g., speaker diarization as a preprocessing step before the ASR) and including them in the session graph. 

\section{Low-latency Speech Processing} \label{sec:LowLatencySpeechProcessing}


The speech processing component receives a stream of audio packets and sends chunks of text (transcript or translation) to the mediator.
For this two steps are run:

\textbf{Input processing:} First, a voice activity detection generates a speech segment that can be extended when new packets of audio arrive. For this we use WebRTC Voice Activity Detector \cite{py-webrtcvad}. Each audio frame (30ms) is classified if it contains speech or not. Then a moving average is calculated. If it exceeds a certain threshold, a new segment is started. New audio is added to this segment until the moving average falls below a certain threshold and the segment ends. 
Second, the backend model (ASR or speech translation (ST)) is run. If there exist speech segments that already ended, they are processed only once and the output is sent as stable text, other segments are constantly processed until they end.

\textbf{Stability detection and output processing:}
We use the method local agreement two (LA2) from \citet{polak2022cuni}. The intuition is that if the prefix of the output stays the same when adding more audio, the prefix should be considered stable. Let $C$ denote the chunk size hyperparameter (\emph{LA2\_chunk\_size}). 
The fixed mode works as follows (see Figure \ref{fig:stability_detection}): It waits until the segment contains
(at least)
$C$ seconds of audio (denoted by $M_1$) and then runs the model but does not output any stable text. Let's denote this first model output by $H_1$. After the segment contains
(at least)
$C$ more seconds of audio (denoted by $M_2$) the model is run again with all the audio and outputs $H_2$. Then the component outputs the common prefix of $H_1$ and $H_2$ as stable output $S_2$. After the segment again contains
(at least)
$C$ more seconds of audio (denoted by $M_3$) the model is run again with all the audio.
However, now $S_2$ is forced as prefix in the ASR/ST model decoding. The model outputs $H_3$ and the common prefix from $H_2$ and $H_3$ is the next stable output $S_3$. This procedure is continued until the speech segment ends.

Note that the ASR/ST model has a certain maximum input size due to latency, memory and compute constraints. Therefore, if this limit is reached, the input audio to the model as well as the corresponding forced prefix is cut away. 

The revision mode differs from the fixed mode in that the last hypothesis except the common prefix is sent as unstable output. Furthermore, in the time period until the speech segment contains again $C$ more seconds of audio, the currently given audio is run through the model and the hypothesis except the last stable output is sent as unstable output.

\section{Low-latency Text Processing} \label{sec:LowLatencyTextProcessing}

The text processing component receives a stream of (potentially revisable) text messages and sends chunks of text (translation) to the mediator.

\textbf{Input processing:} First, all input text that arrived is split into sentences by punctuation. Then, the backend model (MT) is run.

\textbf{Stability detection and output processing:} All sentences containing only stable text are processed once and the output is sent as stable text. For the other sentences containing unstable text the behavior depends on the mode. If text is stable or not is given by the speech processing component.

The revision mode works as follows: All sentences containing unstable text are processed by the backend model and the output text is sent as unstable text. A similar approach is not possible in the speech processing revision mode (see Section \ref{sec:LowLatencySpeechProcessing}) since speech segments are not limited in size but the model input size is.

For the fixed mode we use the method local agreement from \citet{liu20s_interspeech}.
The processing is similar to the speech processing.
The difference is that the backend model is run when at least one new word is given instead of at least $C$ seconds of audio. In our preliminary experiments, up to at least five words but the results were basically identical since the input is extended by a few words most of the time.
Furthermore, only the stable part of the sentences containing unstable text is used as input. This restriction is not necessary in the speech processing component since there is no unstable audio input.

\section{Evaluation Framework}
We evaluate our system in an end-to-end fashion. 
That is, given an input audio, we send it to the system and evaluate the final returned transcript and translation. 
We provide an evaluation framework\footnote{\url{https://git.scc.kit.edu/isl/lt-middleware/lt-evaluation}} that assess the system in different aspects and logs the results to categorized experiments on an UI board using MLflow \cite{Zaharia2018AcceleratingTM}.
We consider different evaluation metrics  as follows.

\textbf{BLEU:} In order to assess the translation quality, we use case-sensitive BLEU score, calculated using sacreBLEU \cite{post-2018-call}. 
We extract the final stable
translation, align it sentence-wise with the gold reference using mwerSegmenter \cite{matusov-etal-2005-evaluating} before calculating the BLEU score. 

\textbf{WER:} In order to assess the transcription quality of the ASR component in the cascaded setting, we use the case-sensitive Word Error Rate (WER) calculated using JiWER\footnote{\url{https://github.com/jitsi/jiwer}}. 
Similar as before, we extract the final stable
transcription, align it sentence-wise with the gold reference using mwerSegmenter \cite{matusov-etal-2005-evaluating} before calculating the WER. 

\textbf{Latency:} We evaluate the latency of the system in an end-to-end manner. Factors such as network latency influence our latency metrics. However, our experiments are conducted locally, thus such factors are constant and negligible.

We define the end-to-end latency of the system as the average time (in seconds) it takes since an utterance is spoken until its first-unchanged translation is returned by the system. 
Note that the first-unchanged translation is not necessarily already marked as ``stable" by the system.

\begin{figure}[t]
  \centering
  \includegraphics[width=1.0\columnwidth]{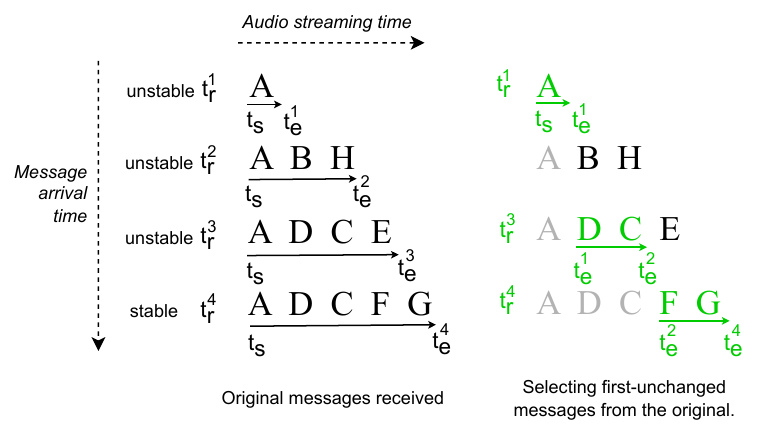}
  \vskip -5pt
  \caption{Example of collecting first-unchanged messages from an unstable-to-stable message block. The first-unchanged messages are in green.}
  \label{fig:extractFirstUnchanged}
  \vskip -10pt
\end{figure}

For each message returned by the system, we have the stable/unstable flag along with three timestamps, $t_s, t_e, t_r$.
The timestamps $t_s$ and $t_e$ are the start and end time of the audio segment that aligns to the message.
The timestamp $t_r$ is when the message was received.
We collect the first unchanged messages as follows. We split the received messages into blocks of messages marked from ``unstable" to ``stable".  
In each unstable-to-stable block, from the last stable message, we backtrack the previously received unstable messages to find the first ones that has prefix-overlaps with the final stable message. The illustration is shown in Figure \ref{fig:extractFirstUnchanged}.

Once we have collected the first-unchanged messages, we can calculate the latency. We use the same definition of delay as \citet{niehues2016dynamic}, where the average delay of the $i^{th}$ message is:
$$
d(t_s^i,t_e^i,t_r^i) = t_r^i - \frac{t_s^i+t_e^i}{2}.
$$

Then we calculate the latency as the weighted average of the delays of all $m$ first-unchanged messages based on their length:
$$
D = \frac{
\sum_{i=1}^{m} d(t_s^i,t_e^i,t_r^i) * (t_e^i - t_s^i)
}{
\sum_{i=1}^{m}(t_e^i - t_s^i)
}.
$$

Note that the timestamps $t_s$ and $t_e$ in our latency formula are calculated by the used streaming algorithm.
Therefore, we also tried another model-independent latency metric that only uses $t_r$. This metric approximates the segment-message alignment by assuming that each word output by the system has the duration of 0.3 second in the audio. Due to the strong assumption, this metric does not represent well the perceived latency. We only use this metric in order to verify our main model-dependent latency metric. 

We find that the model-independent latency metric and our model-dependent metric provide the same relative ranking of the systems. This indicates 
that the timestamps $t_s$ and $t_e$ provided by the model itself are reliable to measure latency.

    


\begin{figure}[t]
  \centering
  \includegraphics[width=0.675\columnwidth]{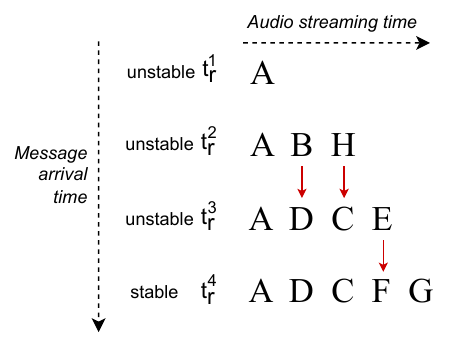}
  \vskip -10pt
  \caption{Example of flickers (denoted by red arrows) in an unstable-to-stable message block.}
  \label{fig:flickers}
  \vskip -10pt
\end{figure}

\textbf{Flickering rate}: The flickering rate is the average number of flickers per reference word. We count the number of flickers by looking at every pair of consecutive messages in a message block. If two words in the same position in the two messages differ, then it is counted as a flicker (see Figure \ref{fig:flickers}). 
The flickering rate is calculated as the total number of flickers divided by the total number of words in the reference.

\section{Experimental setup}
\subsection{Evaluation data}
We test our system using datasets from different language pairs. Test datasets includes:
\begin{itemize}
    \item Test data from the IWSLT shared task (\textit{tst19, tst20}) \cite{anastasopoulos-etal-2021-findings, anastasopoulos-etal-2022-findings}, where the domain is TED talks.
    \item The test split of Multilingual TEDx Corpus (\textit{mTEDx}) \cite{salesky2021multilingual}, where the domain is TED talks.
    \item Lecture data (\textit{LT}) which we collected internally at our university. This test set include a \textit{CS} variance which includes lectures on the Computer Science domain, and a \textit{nonCS} variance which includes lectures outside of the Computer Science domain.
    \item ACL development (\textit{ACL dev}) set \cite{salesky-2023-evaluating}, where the domain is ACL conference talks.
\end{itemize}
The detailed statistics of the test data is shown in Table \ref{tab:testDataStats}.

\begin{table}[t]
\centering

\begin{tabular}{lrrr}
Test data                 & Lang. pair                     & \multicolumn{1}{l}{Hours}                 & \multicolumn{1}{l}{\# Utt.}                 \\ \hline
tst19                     & en$\rightarrow$de                 & 4.82                                      & 2279                                              \\
tst20                     & en$\rightarrow$de                 & 4.09                                      & 1804                                              \\
LT CS                     & de$\rightarrow$en                 & 6.39                                      & 2454                                              \\
LT nonCS                  & de$\rightarrow$en                 & 2.66                                      & 1516                                              \\
mTEDx                     & es$\rightarrow$en                 & 2.07                                      & 1012                                              \\
                          & it$\rightarrow$en                 & 2.16                                      & 999                                               \\
ACL dev *                 & en$\rightarrow$X                  & 0.95                                      & 468                                               \\
\end{tabular}
\vskip -5pt
\caption{Statistics of the test data. *Test data containing en audio with translations into de, ja, zh, ar, nl, fr, fa, pt, ru and tr.}
\label{tab:testDataStats}
\vskip -10pt
\end{table}

\begin{table}[t]
\centering
\resizebox{\columnwidth}{!}{%
\begin{tabular}{llrrrlrr}
        & \multicolumn{1}{l}{} & \multicolumn{3}{c}{Cascaded ST}                                                                 &  & \multicolumn{2}{c}{E2E ST}                                              \\ \cline{3-5} \cline{7-8} 
Testset & C    & W $\downarrow$ & \multicolumn{1}{c}{B $\uparrow$} & \multicolumn{1}{c}{L $\downarrow$} &  & \multicolumn{1}{c}{B $\uparrow$} & \multicolumn{1}{c}{L $\downarrow$} \\ \hline
tst19   & .5                  & 20.8             & 21.6                                & 3.6                                      &  & 20.5                                & 2.1                                      \\
        & 1                    & 17.0             & 24.6                                & 5.6                                      &                      & 22.8                                & 2.6                                      \\
        & 2                    & 16.4             & 25.5                                & 6.8                                      &                      & 23.2                                & 3.9                                      \\
        & 3                    & 16.6             & 25.7                                & 7.8                                      &                      & 23.6                                & 5.0                                      \\ \hline
ACL     & .5                  & 18.7             & 29.8                                & 4.3                                      &                      & 22.4                                & 2.1                                      \\
dev     & 1                    & 16.7             & 32.6                                & 6.2                                      &                      & 25.4                                & 2.7                                      \\
        & 2                    & 16.7             & 34.2                                & 7.4                                      &                      & 26.5                                & 4.0                                      \\
        & 3                    & 17.2             & 35.2                                & 8.6                                     &                      & 26.2                                & 5.3                                     
\end{tabular}
}
\vskip -5pt
\caption{Quality vs. latency (in fixed mode). \mbox{C: \textit{LA2\_chunk\_size}} (s), W: WER, B: BLEU score, \mbox{L: Latency (s) of the translation output}.}
\label{tab:LA2_chunk_size}
\vskip -10pt
\end{table}

\subsection{Transcription and translation models}
The English ASR models are built based on pretrained WavLM~\cite{chen2022wavlm} and BART~\cite{lewis2019bart}\footnote{With the recipe available at \href{https://huggingface.co/nguyenvulebinh/wavlm-bart}{here}.}, while for Multilingual ASR we utilized the XLS-R models~\cite{babu2021xls} for the encoder and the MBART-50 model~\cite{liu2020multilingual} for the decoder following~\cite{pham-etal-2022-effective}. On the other hand, the translation models are based on the pretrained DeltaLM~\cite{ma2021deltalm}. For the en$\rightarrow$X direction, the models are fine-tuned to optimize for ACL talks based on~\citet{liu-etal-2023-kits}. For other directions, DeltaLM is fine-tuned on the combination of commonly available datasets\footnote{Paracrawl, UNPC, EUBookshop, MultiUN, EuroPat, TildeMODEL, DGT, Europarl, QED and NewsCommentary.}.

Finally, for the end-to-end ST system, we used the language-agnostic model from~\citet{huber2022code} that can decode en-de ST and de ASR.


\section{Results and Discussion}
\label{sec:results}

\subsection{Quality vs Latency trade-off}

In the first experiment, we assess the trade-off between translation quality and latency by modifying the $LA2\_chunk\_size$ parameter.
The results are shown in Table \ref{tab:LA2_chunk_size}. As can be seen, as we increase chunk size, the translation quality improves while the latency gets worse, both for cascaded ST and end-to-end ST. This is expected, since higher chunk size means longer input given to the model at each step, thus the output has better quality due to having more context, while the latency gets worse due to more waiting time for collecting the input.


\subsection{Revision mode vs fixed mode}

\begin{figure}[t]
  \centering
  \includegraphics[width=\columnwidth]{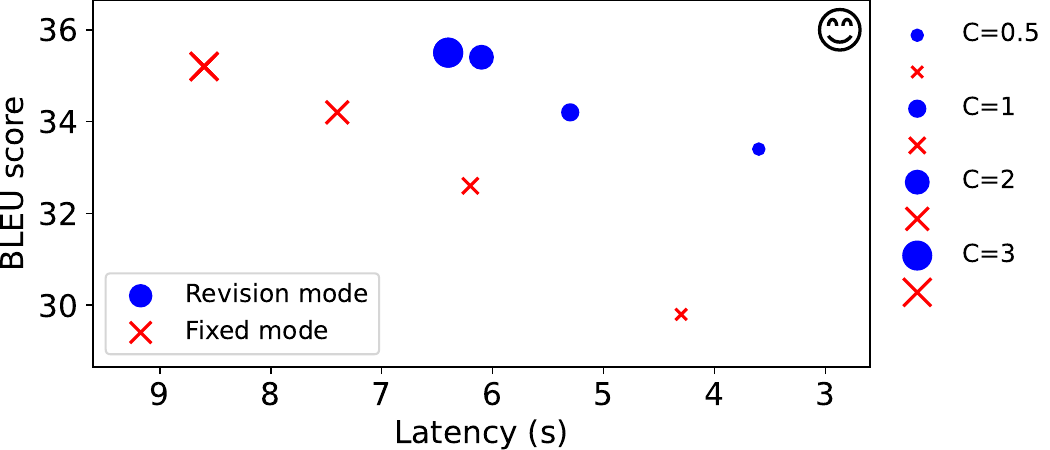}
  \vskip -5pt
  \caption{Latency vs. quality (for the cascaded model) in revision mode or fixed mode. \mbox{C: $LA2\_chunk\_size$ (s)}.}
  \label{fig:revision_vs_fixed_mode}
  \vskip -10pt
\end{figure}

Second, we report the results of comparing the revision mode to the fixed mode with different $LA2\_chunk\_size$ values when performing cascaded translation on the en-de ACL dev set. As can be seen in Figure \ref{fig:revision_vs_fixed_mode}, in general, revision mode has better BLEU score and better latency than fixed mode.
The better BLEU score comes from the revision mode being able to correct its previous output when more audio input is provided as time goes by.

\subsection{Cascaded vs End-to-End}

\begin{figure}[t]
  \centering
  \includegraphics[width=\columnwidth]{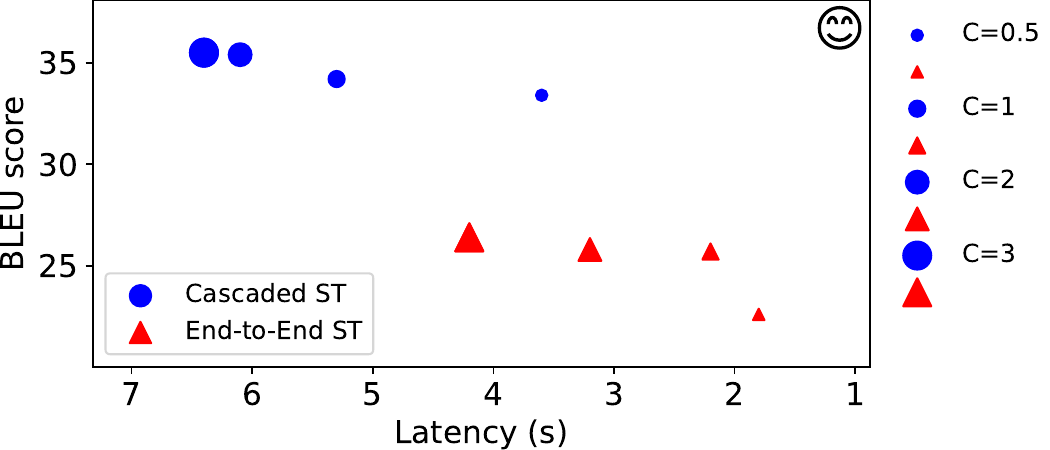}
  \vskip -5pt
  \caption{Latency vs. quality (in revision mode) for the cascaded ST or End-to-End ST model. \mbox{C: $LA2\_chunk\_size$} (s).}
  \label{fig:cascaded_vs_e2e}
  \vskip -10pt
\end{figure}

Third, we report the results of comparing the cascaded setting to the end-to-end setting when performing online translation with revision mode on the ACL dev set. As can be seen in Figure \ref{fig:cascaded_vs_e2e}, in general, cascaded ST has better BLEU score yet worse latency than end-to-end ST. Cascaded ST has worse latency since it contains two components and each component has to do computation.
However, we observe that, with a similar latency of $\sim$ 3.5 seconds, cascaded ST still obtains a better BLEU score. On the other hand, end-to-end ST has a better minimum latency that can be achieved (almost two seconds lower than the cascaded system).

\subsection{Load balancing}

In order to assess the system's capability to balance loads, we conduct experiments on running multiple sessions simultaneously using the same hosted model, with and without scaling the system's number of middleware workers.
For speech processing, we test parallel sessions on ACL dev en-de using the end-to-end ST model.
For text processing, we test one cascaded ST session on ACL dev where the number of parallel sessions is the number of requested MT languages. In all experiments, we set $LA2\_chunk\_size=2$. We report only the en-de results.

The results are shown in Table \ref{tab:parallel}. As expected, the latency gets worse as the number of parallel sessions increases. Using multiple middleware workers counteracts that to some extent by making sure that the backend model is always busy and not waiting for the next request.
Furthermore, we see that when the number of parallel sessions increases, the flickering rate decreases. This is because during higher load, fewer requests are sent to the backend and we observe less flickering. Here our automatic load balancing can be seen in action.

\begin{table}[t]
\resizebox{\columnwidth}{!}{%
\begin{tabular}{rrrrrrrrrr}
          & \multicolumn{1}{l}{} & \multicolumn{3}{c}{Speech processing}                              &  & \multicolumn{3}{c}{Text processing}                                 \\ \cline{3-5} \cline{7-9} 
          w & s               & B $\uparrow$ & L $\downarrow$ & F $\downarrow$ &  & B $\uparrow$ & L $\downarrow$ & F $\downarrow$ \\ \hline
1  & 1                    & 25.9            & 3.2                  & 0.5                          &  & 34.9            & 6.7                 & 0.5                          \\
           & 2                    & 26.1            & 20.2                  & 0.5                          &  & 34.6            & 8.4                   & 0.4                          \\
          & 5                    & 21.3            & 28.2                 & 0.2                          &  & 34.8            & 28.1                 & 0.2                          \\ \hline
5 & 1                    & 26.2            & 3.2                  & 0.6                          &  & 35.1            & 6.1                 & 0.5                          \\
          & 2                    & 26.5            & 4.6                  & 0.5                          &  & 33.6            & 8.0                 & 0.5                          \\
          & 5                    & 25.3            & 16.7                 & 0.3                          &  & 34.5            & 15.9                 & 0.2                         
 
\\
\end{tabular}
}
\vskip -5pt
\caption{Quality, latency and flickering rate when scaling the number of sessions (with one hosted model per language). w: number of middleware workers, \mbox{s: number of parallel sessions}, B: Quality (BLEU score), \mbox{L: Latency} (s), F: Flickering rate. $LA2\_chunk\_size$ is set to 2 seconds.}
\label{tab:parallel}
\vskip -10pt
\end{table}

\section{Related work}
SimulEval \citep{ma-etal-2020-simuleval} provides an evaluation framework for low-latency simultaneous speech translation with a decoupled client-server architecture allowing to plug-in translation models and stability detection policies.
As the main difference we leave the audio segmentation up to the model whereas \citet{ma-etal-2020-simuleval} rely on a pre-segmentation of the audio, we factor in the computational latency in addition to the model latency and explore the scaling behavior in multi-session scenarios, both for a more realistic deployment scenario.
Similar to this work \citet{franceschini2020removing} implement a low-latency speech translation pipeline, however, their architecture does not scale well to multiple sessions and is not well suited for end-to-end evaluation.


\section{Limitations and Conclusion}
\label{sec:bibtex}


Since we run and evaluate the experiments in a realistic real-world scenario, it is difficult to exactly reproduce the results. The experiments are non-deterministic, e.g., because of network latencies. Furthermore, the results depend on the speed of the used hardware, especially the used hardware for the backend models.
Additionally, we expect that each streaming algorithm implemented returns start and end timestamps. This may not be the case for all streaming algorithms one could want to compare.

In conclusion, this paper presented a framework for running and evaluating low-latency speech translation under realistic conditions. The research opens up new possibilities for advancing low-latency translation systems and serves as a resource for researchers seeking to improve the latency and quality of real-time speech translation applications by being able to properly evaluate different models and streaming algorithms.




\section*{Acknowledgements}

The projects on which this paper is based were funded by the Federal Ministry of Education and Research (BMBF) of Germany under the numbers 01IS18040A (OML) and 01EF1803B (RELATER).

\bibliography{anthology,custom}
\bibliographystyle{acl_natbib}

\appendix

\section{Detailed results} \label{appendix:DetailedResults}

\begin{table*}[htbp]
\resizebox{\textwidth}{!}{%
\begin{tabular}{llrrrrrrrr}
                                                                                       &       & \multicolumn{1}{c}{Offline}         & \multicolumn{1}{c}{} & \multicolumn{3}{c}{Online: Revision mode}                                                                                                 & \multicolumn{1}{c}{} & \multicolumn{2}{c}{Online: Fixed mode}                                                 \\ \cline{3-3} \cline{5-7} \cline{9-10} 
                                                                                       &       & \multicolumn{1}{c}{BLEU $\uparrow$} & \multicolumn{1}{c}{} & \multicolumn{1}{c}{$\Delta$BLEU $\uparrow$} & \multicolumn{1}{c}{Latency $\downarrow$} & \multicolumn{1}{c}{Flickering rate $\downarrow$} & \multicolumn{1}{c}{} & \multicolumn{1}{c}{$\Delta$BLEU $\uparrow$} & \multicolumn{1}{c}{Latency $\downarrow$} \\ \hline
\multirow{2}{*}{\begin{tabular}[c]{@{}l@{}}TED\\ (en$\rightarrow$de)\end{tabular}}     & tst19 & 27.2                                &                      & -1.3                                        & 5.4                                      & 0.5                                              &                      & -1.7                                        & 5.3                                      \\
                                                                                       & tst20 & 29.8                                &                      & -1.1                                        & 5.1                                      & 0.5                                              &                      & -1.5                                        & 6.9                                      \\ \hline
\multirow{2}{*}{\begin{tabular}[c]{@{}l@{}}LT\\ (de$\rightarrow$en)\end{tabular}}      & CS    & 25.2                                &                      & -2.0                                        & 5.6                                      & 0.6                                              &                      & -2.4                                        & 6.0                                      \\
                                                                                       & nonCS & 28.5                                &                      & -0.2                                        & 7.1                                      & 0.6                                              &                      & -1.2                                        & 5.7                                      \\ \hline
\multirow{2}{*}{\begin{tabular}[c]{@{}l@{}}mTEDx\\ (X$\rightarrow$en)\end{tabular}}    & es    & 31.0                                &                      & -2.5                                        & 7.5                                      & 0.4                                              &                      & -2.6                                        & 7.6                                      \\
                                                                                       & it    & 31.5                                &                      & -4.2                                        & 12.5                                     & 0.5                                              &                      & -5.1                                        & 11.6                                     \\ \hline
\multirow{2}{*}{\begin{tabular}[c]{@{}l@{}}ACL dev \\ (en$\rightarrow$X)\end{tabular}} & de    & 36.5                                &                      & -1.9                                        & 6.6                                      & 0.5                                              &                      & -2.0                                        & 7.5                                      \\
                                                                                       & ja    & 39.6                                &                      & -1.6                                        & 8.4                                      & 0.1                                              &                      & -5.2                                        & 8.7                                      \\
                                                                                       & zh    & 45.3                                &                      & -0.5                                        & 8.0                                      & 0.1                                              &                      & -4.6                                        & 8.0                                      \\
                                                                                       & ar    & 28.3                                &                      & -0.8                                        & 6.8                                      & 0.5                                              &                      & -1.1                                        & 7.3                                      \\
                                                                                       & nl    & 42.7                                &                      & -1.3                                        & 6.4                                      & 0.5                                              &                      & -2.5                                        & 7.4                                      \\
                                                                                       & fr    & 43.7                                &                      & -0.4                                        & 5.9                                      & 0.5                                              &                      & -1.0                                        & 7.6                                      \\
                                                                                       & fa    & 21.8                                &                      & -0.8                                        & 6.8                                      & 0.7                                              &                      & -1.8                                        & 7.5                                      \\
                                                                                       & pt    & 45.2                                &                      & -1.2                                        & 6.0                                      & 0.4                                              &                      & -1.8                                        & 7.3                                      \\
                                                                                       & ru    & 13.5                                &                      & -1.0                                        & 6.5                                      & 0.5                                              &                      & -1.2                                        & 7.4                                      \\
                                                                                       & tr    & 20.1                                &                      & -0.9                                        & 6.6                                      & 0.5                                              &                      & -1.2                                        & 7.4                                      \\
                                                                                                                              
\end{tabular}
}
\vskip -0pt
\caption{Overall performance of our cascaded system with $LA2\_chunk\_size$ set to 2 seconds: Quality, latency and flickering rate. $\Delta$BLEU: difference compared the corresponding offline setting.}
\label{tab:overallPerformance}
\end{table*}

We report the overall performance of our system on different test data and language pairs with different settings at Table \ref{tab:overallPerformance}. In this experiment, we use the cascaded setting with $LA2\_chunk\_size=2$. 
As can be seen, the BLEU scores drop around by one point when we move from offline to online setting (in fixed mode a little more), depending on the language directions. 


\section{Additional information}

A video demonstrating the system can be found here: \href{https://lt2srv.iar.kit.edu/archive/%252F%252FOther%252FEMNLP2023}{Video link}


\end{document}